%% file: main.tex
\documentclass{article} 
\input{preprint_layout.tex}
\usepackage{times}
\usepackage{graphicx}

\input{math_commands.tex}

\usepackage{hyperref}
\usepackage{url}

\title{Unravel: Recovering an Interactive 3D Scene from a Single Image}
\vspace{12pt} 

\author{%
\parbox{0.95\textwidth}{%
\centering
\normalfont\small
\makebox[\linewidth][s]{%
\mbox{Hui Shan$^{1,2,3}$}\hfill
\mbox{Keyang Luo$^{4}$}\hfill
\mbox{Binjia Zhou$^{1}$}\hfill
\mbox{Ming Li$^{1,2,3}$}\hfill
\mbox{Kai Li$^{7}$}\hfill
\mbox{Xin Wang$^{3}$}\hfill
\mbox{Zhenyang Li$^{8}$}%
}\\[3pt]
\makebox[\linewidth][s]{%
\mbox{Zesong Yang$^{1}$}\hfill
\mbox{Weikai Chen$^{4}$}\hfill
\mbox{Runze Zhang$^{4}$}\hfill
\mbox{Xin WANG$^{4}$}\hfill
\mbox{Yanwei Fu$^{6}$}\hfill
\mbox{Zhen Chen$^{5}$}\hfill
\mbox{Xiangru Huang$^{3}$\thanks{Corresponding author:
\href{mailto:huangxiangru@westlake.edu.cn}{\texttt{huangxiangru@westlake.edu.cn}}}}%
}\\[7pt]
\normalfont\normalsize
\makebox[\linewidth][s]{%
\mbox{$^{1}$Zhejiang University}\hfill
\mbox{$^{2}$Shanghai Innovation Institute}\hfill
\mbox{$^{3}$Westlake University}\hfill
\mbox{$^{4}$LIGHTSPEED}%
}\\[3pt]
\makebox[\linewidth][s]{%
\mbox{$^{5}$Adobe}\hfill
\mbox{$^{6}$Fudan University}\hfill
\mbox{$^{7}$City University of Hong Kong}\hfill
\mbox{$^{8}$University of Hong Kong}%
}%
}%
}

\definecolor{mblue}{rgb}{0.40, 0.67, 0.88}

\usepackage{booktabs}
\begin{document}

\maketitle
\vspace{-10pt}

\begin{figure}[h]
  \centering
  \includegraphics[width=0.9\linewidth]{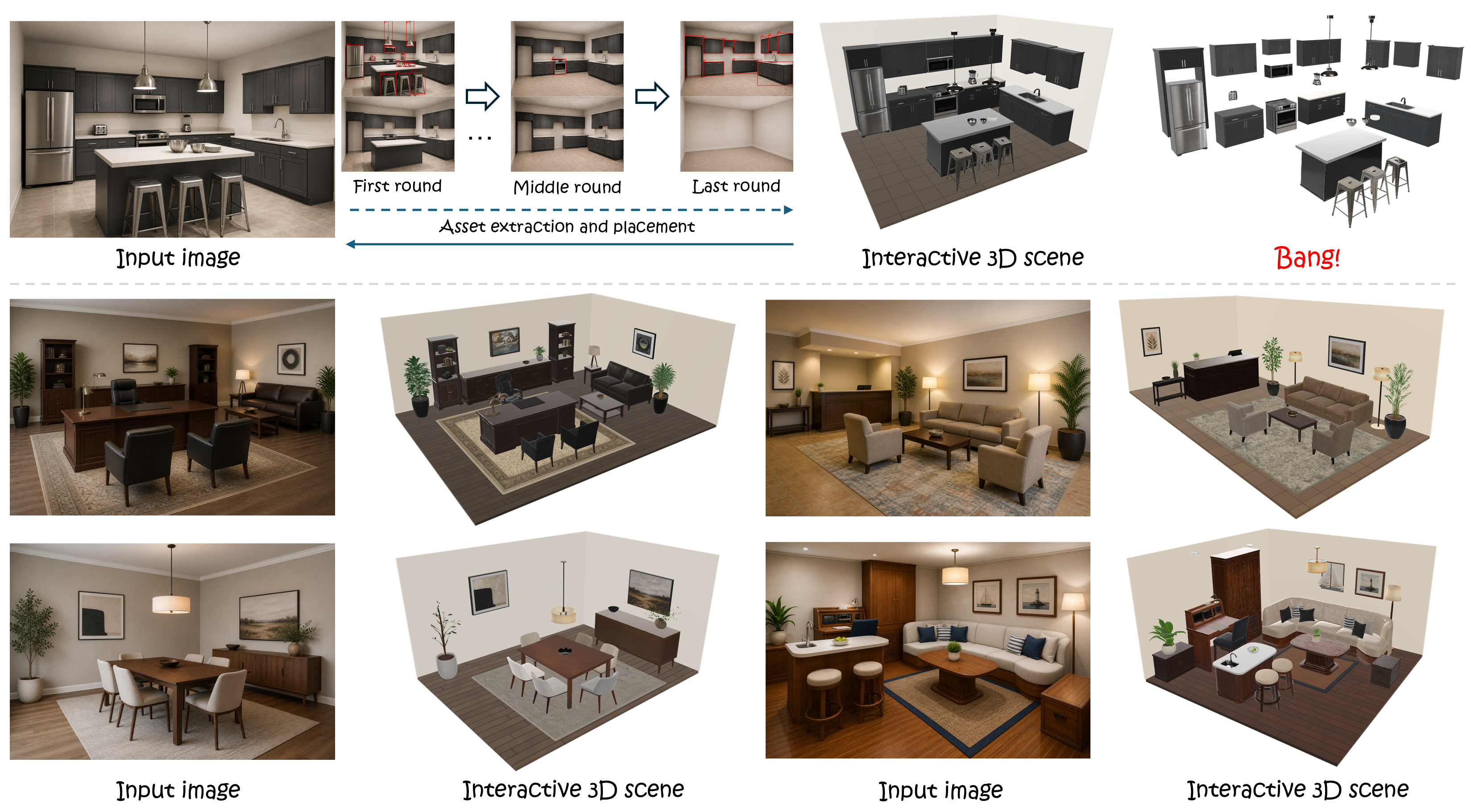}
  \caption{Unravel turns a single image into an interactive 3D scene. In the top row, objects are removed from the input image layer by layer until the room is empty, and the scene is then rebuilt with every object as an independent asset (exploded view, right). The rows below show more input images and their recovered interactive scenes.}
  \label{fig:teaser}
\end{figure}

\input{Section/0.abstract}
\input{Section/1.Intro}

\input{Section/2.RelatedWork}
\input{Section/3.Methods}
\input{Section/4.Experiments}

\input{Section/5.Conclusions}

\bibliography{references}
\bibliographystyle{references}

\appendix
\section{Appendix}
\subsection{Accurate and Stable Bounding Box Localization}
\label{appendix:boxes}

Every removal target needs an accurate box, since the editor removes exactly what the boxes mark. Each target comes with a name and a short description. A vision-language model boxes all targets in one call, one box per index, so every box stays bound to its target. The prompt requires tight boxes that touch the outermost pixels of the object. Two boxes with IoU of at least $\tau_{dup}$ are treated as one object boxed twice, and the extra one is detected again.

A global call can drift systematically in cluttered scenes, and it cannot notice its own drift. Each global box is therefore probed with one independent local sample of the same target. If the probe box overlaps the global box with IoU at least $\tau_p$, the global box is confirmed. Otherwise the box is considered drifted, and the target is re-detected by local detection. Local detection samples the VLM $n$ times independently for that target and keeps the samples that agree with at least one other sample,

\begin{equation*}
    \mathcal{C} \;=\; \{\, b_i \;:\; \exists\, j \neq i,\ \mathrm{IoU}(b_i, b_j) \ge \tau_c \,\}.
\end{equation*}

The element-wise median of these samples suppresses outliers and replaces the drifted box. Whenever the probe or local detection fails to return a box, the global box is kept. The global call covers all targets at once, while local detection stays stable. The cross-check keeps the strengths of both. We use $n = 3$, $\tau_{dup} = 0.85$, $\tau_c = 0.7$, and $\tau_p = 0.5$ in all experiments.

The verified target boxes are drawn in red with their indices, and the objects to protect in green, on the plan image. The annotated image is fed back to the VLM once more, and any movable object still without a box is added as an object to protect. A missed object is therefore protected rather than deleted by mistake.

\subsection{Registration and Settling Details}
\label{appendix:registration}

The registration transform combines a global scale with bounded per-axis adjustments in the object's frame. An under-observed axis may only grow, since its measured span underestimates the true one. A well-observed axis may also contract, since its span is reliable. When all three axes agree that the asset is too small, we apply a single uniform scale instead, so the asset keeps its proportions. Each correction is kept only if it leaves the fit $D_i$ no worse, so an asset may grow to cover its points but never shrink away from them. Image cues can refine orientation and translation, but only when this 3D fit is preserved.

An object is supported when its bottom rests on a surface beneath it, either the floor or the top of another object, as a cup rests on a tabletop. Formally, let $\operatorname{bot}(a_i')$ be the height of the lowest point of the settled asset, let $h_{\mathrm{floor}}$ be the floor height, and let $\operatorname{top}_j(a_i')$ be the height of the top surface of another asset $a_j'$ within the footprint of $a_i'$. We set $\operatorname{supported}(a_i') = 1$ when the highest such surface at or below the bottom lies within a small contact tolerance $\epsilon_s$ of it, that is, when
\begin{equation*}
    0 \le \operatorname{bot}(a_i') - \max H_i \le \epsilon_s, \quad H_i = \{h_{\mathrm{floor}}\} \cup \{ \, \operatorname{top}_j(a_i') \mid j \neq i,\ \operatorname{top}_j(a_i') \le \operatorname{bot}(a_i') \, \}.
\end{equation*}
The floor is always in $H_i$, so an object with nothing beneath it is supported exactly when it touches the floor. We use $\epsilon_s = 3\,$cm in all experiments.

The clearance $\phi(a_i', a_j')$ between two assets turns negative as soon as they overlap. The budget $\mathcal{B}_i$ limits how far an asset may move from where it was observed. Let $d(a_i, C_i)$ be the distance between the center of the asset and the center of its cluster before settling. We require
\begin{equation*}
    \mathcal{B}_i=\big\{\delta\in\mathbb{R}^3 \,\big|\, d(a_i', C_i)\le \max\big(d(a_i, C_i),\,\epsilon\big)\big\},
\end{equation*}
where $\epsilon$ is a small tolerance. An asset that starts at its observation may move by at most $\epsilon$, and an asset that starts misplaced may move toward the observation but never further away. We set $\epsilon$ to $0.3\,$m in the horizontal plane and $0.15\,$m along the vertical direction in all experiments.

The validity mask $M_k$ in Section~\ref{subsubsection:driftcontrol} uses a fixed threshold $\tau = 0.5$.

\subsection{Asset Scoring Protocol}
\label{appendix:scoring}

The asset comparison in Section~\ref{subsection:Ablation} is scored as follows. We evaluate all 160 objects from the five test scenes. The human evaluation involved 15 participants. Each object contributes two meshes, one from direct reconstruction and one from Unravel. Both meshes are rendered from the same four viewpoints at $90^\circ$ intervals. The two renders appear side by side in random order, and the rater does not know which mesh comes from which pipeline. Each pair is judged against the same reference, a crop of the object's region from the image of its best-observed round. The rater answers one question, how faithfully each mesh reproduces the object in the reference.

\begin{figure}[t]
  \centering
  \includegraphics[width=\linewidth]{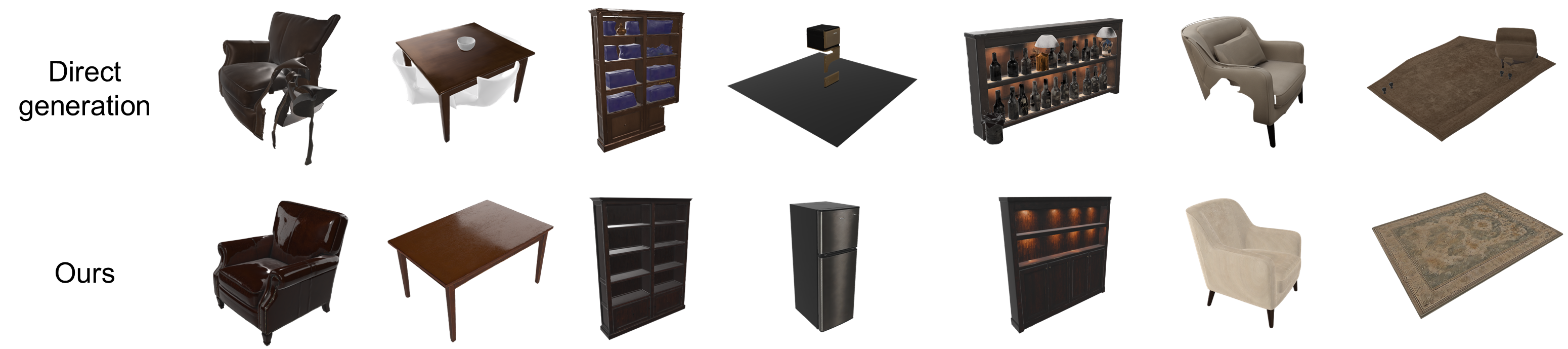}
  \caption{Generated assets with and without deconstruction. Each column shows one object. The top row is generated from the mask in the input view, and the bottom row from the mask captured at the best-observed moment. Incomplete masks lead to broken geometry, spurious parts, and hallucinated contents, while Unravel keeps the assets faithful.}
  \label{fig:asset_ablation}
\end{figure}

Scores range from 1 to 10. A score of 1 to 3 means the shape is wrong or essential parts are missing. A score of 4 to 6 means the object category is recognizable but the structure is deformed or incomplete. A score of 7 or 8 means the overall shape is right with minor surface artifacts. A score of 9 or 10 means the mesh is faithful in shape and structure, including the parts that were occluded in the input image. We use GPT-5.5~\citep{openai_gpt} as the VLM rater. It receives the same renders, reference, and rubric. The model and the human raters score independently, and neither sees the other's ratings. Table~\ref{tab:ablation} reports the mean score of each side over all 160 objects, and Fig.~\ref{fig:asset_ablation} shows representative examples of the quality gap.

\subsection{Additional Scene Reconstruction Results}
\label{appendix:results}

\begin{figure}[t]
  \centering
  \includegraphics[width=\linewidth]{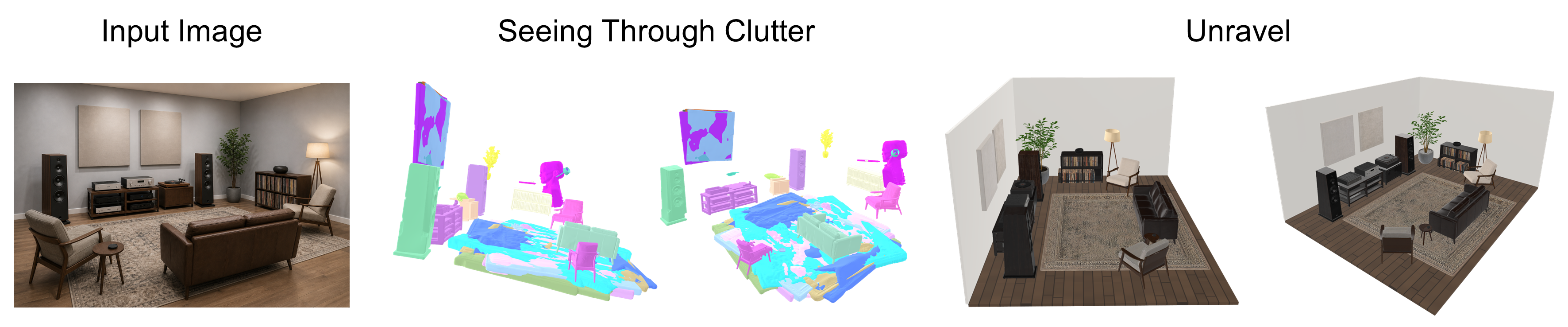}
  \caption{Comparison with Seeing Through Clutter. The input image is on the left, followed by two views of Seeing Through Clutter in the middle and two views of Unravel on the right. Seeing Through Clutter is shown as colored geometry and Unravel as a textured scene.}
  \label{fig:seeing_comparison}
\end{figure}

Figure~\ref{fig:seeing_comparison} compares Unravel with Seeing Through Clutter~\citep{aguina2026seeing} on a cluttered image. We follow its open-source configuration, except that we upgrade its planner to GPT-5.5 to match Unravel. In this example, Seeing Through Clutter produces fragmented geometry and misplaced objects. Unravel more closely preserves the main furniture arrangement.

\begin{figure}[t]
  \centering
  \includegraphics[width=0.85\linewidth]{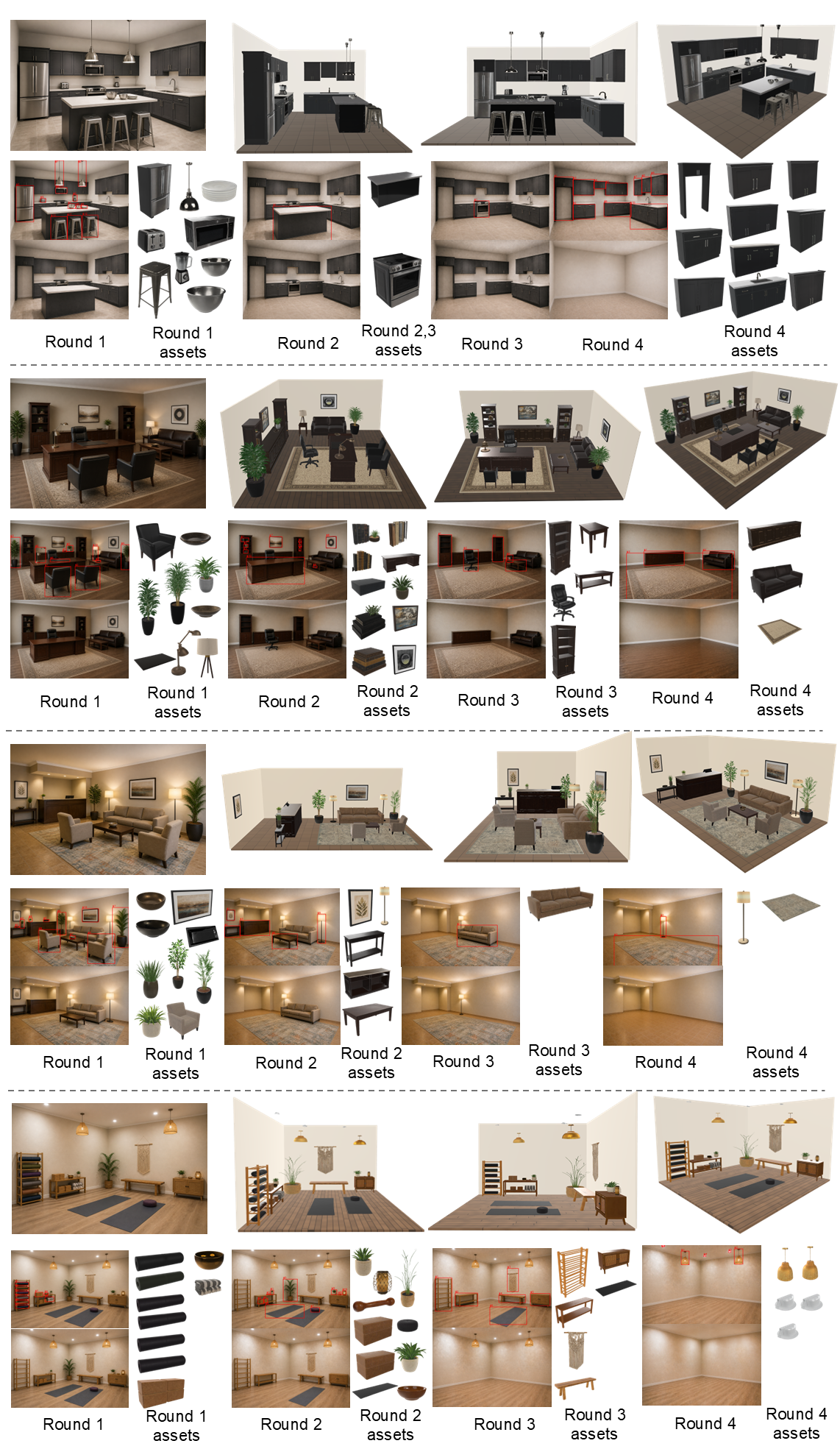}
  \caption{Complete deconstruction traces on four scenes. For each scene we show the input image, three views rendered from the reconstructed scene, the edited image of every round, and the assets extracted in each round. Red boxes mark the removal targets of each round, and duplicate assets are omitted for space.}
  \label{fig:results_1}
\end{figure}

\begin{figure}[t]
  \centering
  \includegraphics[width=0.9\linewidth]{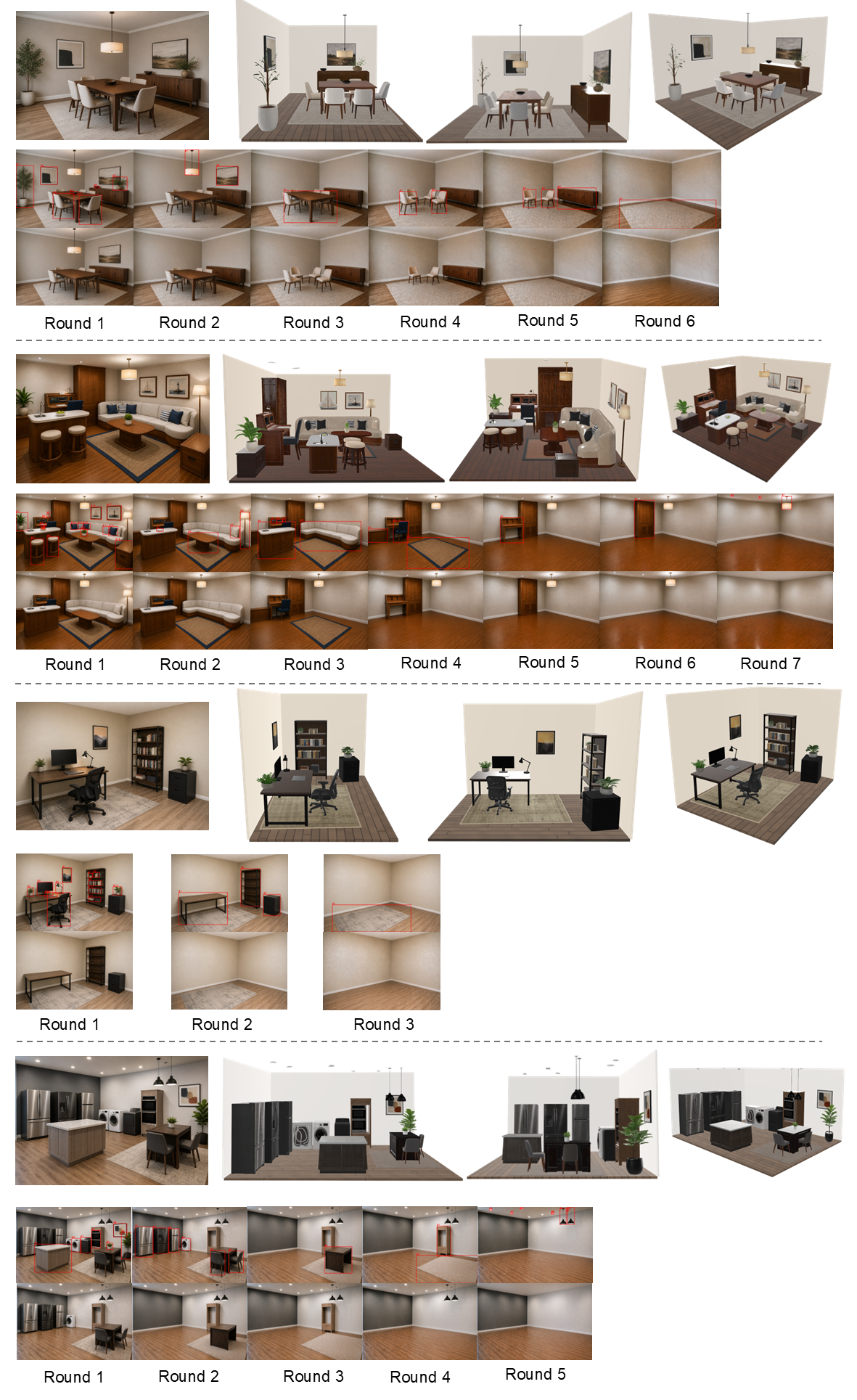}
  \caption{Deconstruction traces on four more scenes. For each scene we show the input image, three views rendered from the reconstructed scene, and the edited image of every round. Red boxes mark the removal targets of each round.}
  \label{fig:results_2}
\end{figure}

\begin{figure}[t]
  \centering
  \includegraphics[width=\linewidth]{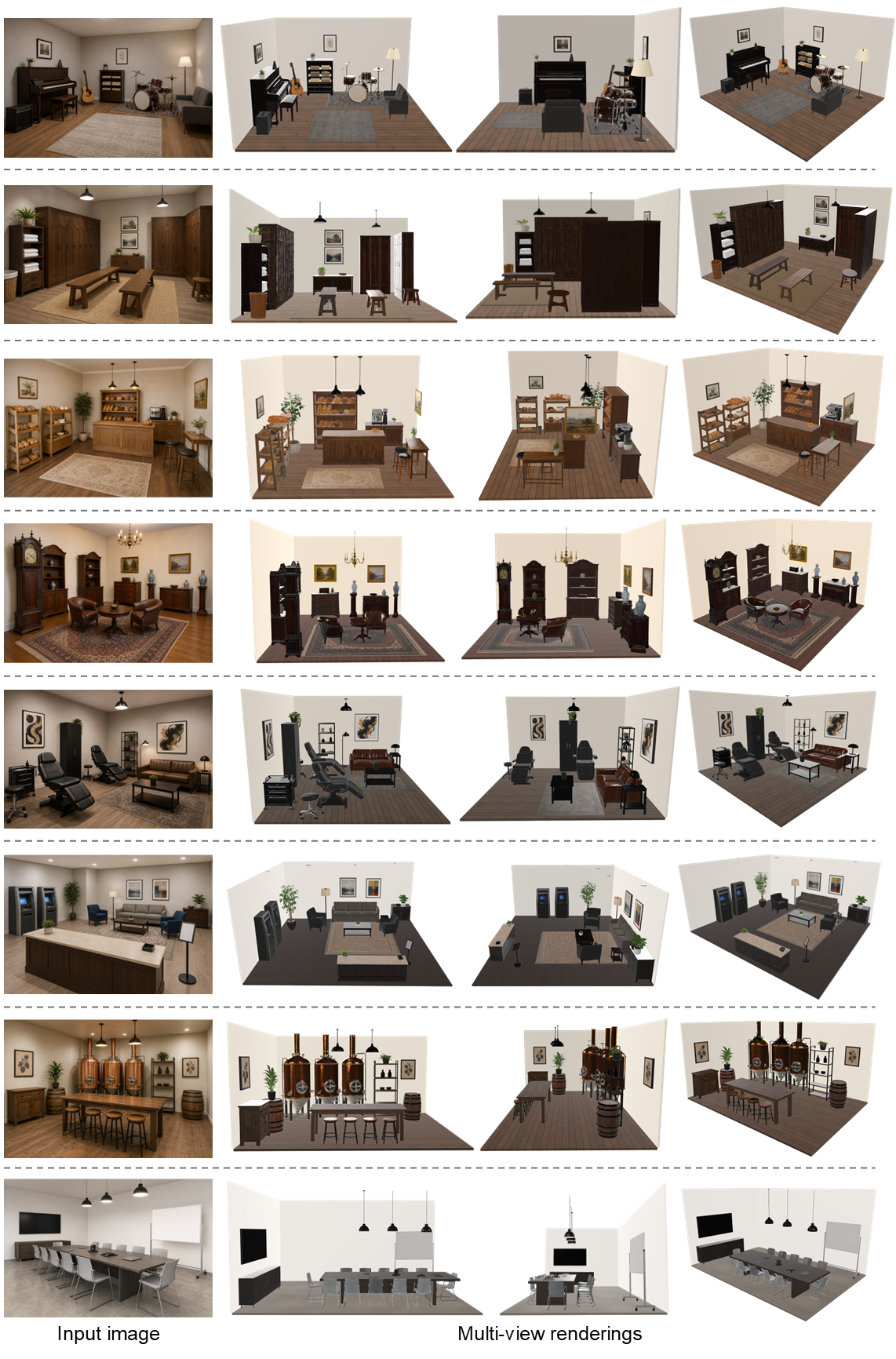}
  \caption{Additional reconstructions on eight further scenes. For each scene we show the input image and three views rendered from the reconstructed scene.}
  \label{fig:results_3}
\end{figure}

Figures~\ref{fig:results_1}, \ref{fig:results_2}, and \ref{fig:results_3} present sixteen additional scenes reconstructed by the same pipeline with identical settings. The scenes cover kitchens, living rooms, and offices, as well as specialty spaces such as a yoga studio, a music room, a bar, and a conference room. Figure~\ref{fig:results_1} traces the full loop and shows the assets extracted in every round. Figure~\ref{fig:results_2} follows the same loop on four more scenes and highlights how the edited image evolves round by round. Figure~\ref{fig:results_3} presents eight further scenes with their input images and multi-view renderings. Across all scenes, the planner removes objects from front to back, the editor restores the freed regions plausibly, and the reconstructed scenes preserve the input layout. These results demonstrate the feasibility and robustness of our pipeline.

\begin{figure}[t]
  \centering
  \includegraphics[width=\linewidth]{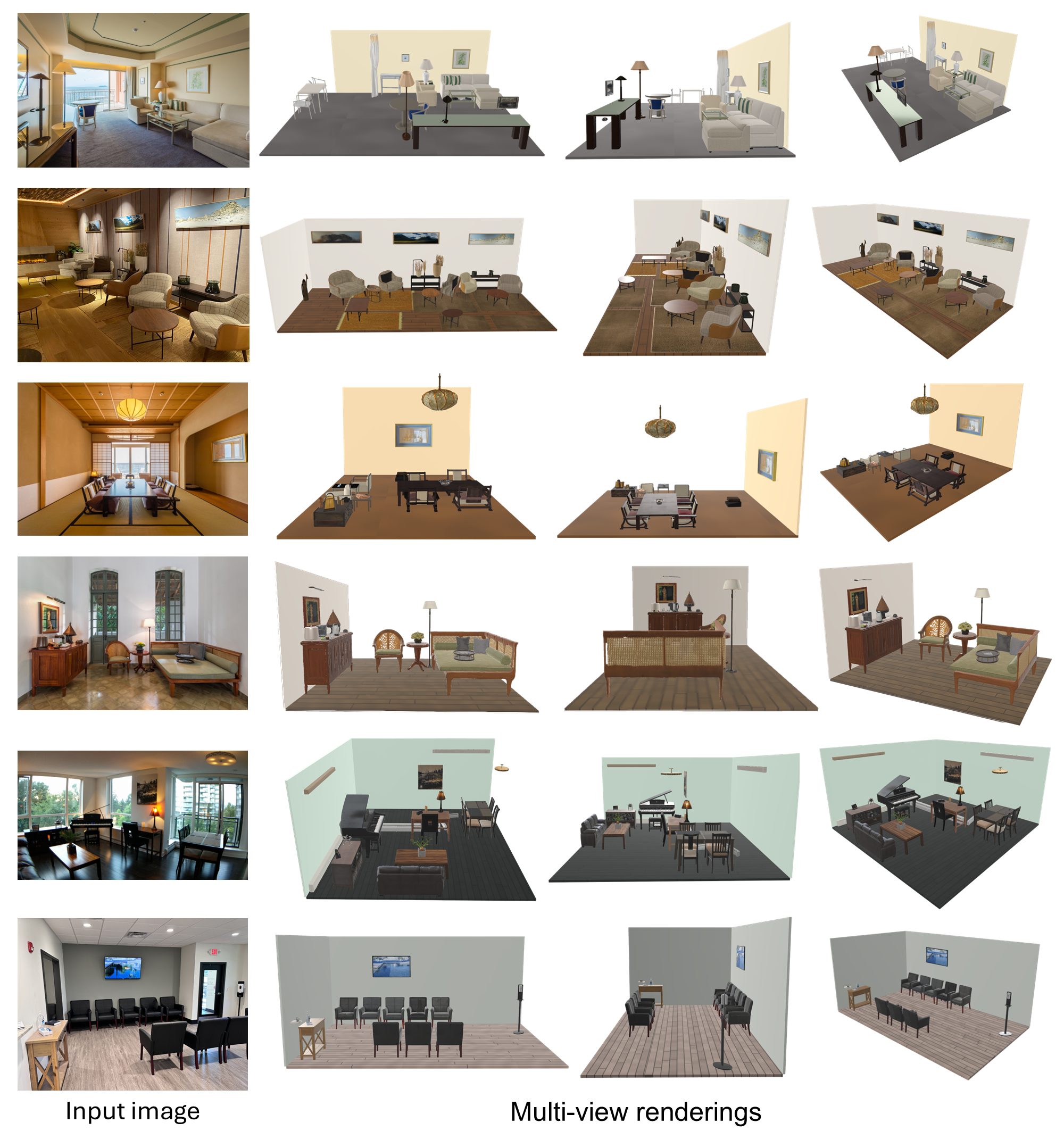}
  \caption{Reconstructions from real photographs. Each row shows an input image on the left and three views of the reconstructed scene on the right.}
  \label{fig:real_results}
\end{figure}

We also present reconstructions from six real indoor photographs in Fig.~\ref{fig:real_results}. The photographs were selected from Wikimedia Commons and show hotel interiors, a residential living room, and a clinic waiting room. The renderings retain the main furniture arrangements.

\end{document}

%% file: preprint_layout.tex
\makeatletter
\RequirePackage{color}
\RequirePackage{fancyhdr}
\RequirePackage{natbib}

\setcitestyle{authoryear,round,citesep={;},aysep={,},yysep={;}}

\def\addcontentsline#1#2#3{}

\def\maketitle{\par
\begingroup
   \def\thefootnote{\fnsymbol{footnote}}
   \def\@makefnmark{\hbox to 0pt{$^{\@thefnmark}$\hss}} %
   \long\def\@makefntext##1{\parindent 1em\noindent
                            \hbox to1.8em{\hss $\m@th ^{\@thefnmark}$}##1}
   \@maketitle \@thanks
\endgroup
\setcounter{footnote}{0}
\let\maketitle\relax \let\@maketitle\relax
\gdef\@thanks{}\gdef\@author{}\gdef\@title{}\let\thanks\relax}

\usepackage{fancyhdr}
\def\@maketitle{\vbox{\hsize\textwidth
{\LARGE\sc \@title\par}

    \def\And{\end{tabular}\hfil\linebreak[0]\hfil
            \begin{tabular}[t]{l}\bf\rule{\z@}{24pt}\ignorespaces}%
  \def\AND{\end{tabular}\hfil\linebreak[4]\hfil
            \begin{tabular}[t]{l}\bf\rule{\z@}{24pt}\ignorespaces}%
    \begin{tabular}[t]{l}\bf\rule{\z@}{24pt}\@author\end{tabular}%

\vskip 0.3in minus 0.1in}}

\renewenvironment{abstract}{\vskip.075in\centerline{\large\sc
Abstract}\vspace{0.5ex}\begin{quote}}{\par\end{quote}\vskip 1ex}

\def\section{\@startsection {section}{1}{\z@}{-2.0ex plus
    -0.5ex minus -.2ex}{1.5ex plus 0.3ex
minus0.2ex}{\large\sc\raggedright}}

\def\subsection{\@startsection{subsection}{2}{\z@}{-1.8ex plus
-0.5ex minus -.2ex}{0.8ex plus .2ex}{\normalsize\sc\raggedright}}
\def\subsubsection{\@startsection{subsubsection}{3}{\z@}{-1.5ex
plus      -0.5ex minus -.2ex}{0.5ex plus
.2ex}{\normalsize\sc\raggedright}}
\def\paragraph{\@startsection{paragraph}{4}{\z@}{1.5ex plus
0.5ex minus .2ex}{-1em}{\normalsize\bf}}
\def\subparagraph{\@startsection{subparagraph}{5}{\z@}{1.5ex plus
  0.5ex minus .2ex}{-1em}{\normalsize\sc}}

\skip\footins 9pt plus 4pt minus 2pt
\def\footnoterule{\kern-3pt \hrule width 12pc \kern 2.6pt }
\leftmargini\leftmargin \leftmarginii 2em
\def\@listi{\leftmargin\leftmargini}
\def\@listii{\leftmargin\leftmarginii
   \labelwidth\leftmarginii\advance\labelwidth-\labelsep
   \topsep 2pt plus 1pt minus 0.5pt
   \parsep 1pt plus 0.5pt minus 0.5pt
   \itemsep \parsep}
\def\@listiii{\leftmargin\leftmarginiii
    \labelwidth\leftmarginiii\advance\labelwidth-\labelsep
    \topsep 1pt plus 0.5pt minus 0.5pt
    \parsep \z@ \partopsep 0.5pt plus 0pt minus 0.5pt
    \itemsep \topsep}
\def\@listiv{\leftmargin\leftmarginiv
     \labelwidth\leftmarginiv\advance\labelwidth-\labelsep}
\def\@listv{\leftmargin\leftmarginv
     \labelwidth\leftmarginv\advance\labelwidth-\labelsep}
\def\@listvi{\leftmargin\leftmarginvi
     \labelwidth\leftmarginvi\advance\labelwidth-\labelsep}

\belowdisplayskip \abovedisplayskip
\def\normalsize{\@setsize\normalsize{11pt}\xpt\@xpt}
\def\small{\@setsize\small{10pt}\ixpt\@ixpt}
\def\footnotesize{\@setsize\footnotesize{10pt}\ixpt\@ixpt}
\def\scriptsize{\@setsize\scriptsize{8pt}\viipt\@viipt}
\def\tiny{\@setsize\tiny{7pt}\vipt\@vipt}
\def\large{\@setsize\large{14pt}\xiipt\@xiipt}
\def\Large{\@setsize\Large{16pt}\xivpt\@xivpt}
\def\LARGE{\@setsize\LARGE{20pt}\xviipt\@xviipt}
\def\huge{\@setsize\huge{23pt}\xxpt\@xxpt}
\def\Huge{\@setsize\Huge{28pt}\xxvpt\@xxvpt}

\makeatother

%% file: math_commands.tex
\usepackage{amsmath,amsfonts,bm}

\def\eqref#1{equation~\ref{#1}}

\def\1{\bm{1}}

\DeclareMathAlphabet{\mathsfit}{\encodingdefault}{\sfdefault}{m}{sl}
\SetMathAlphabet{\mathsfit}{bold}{\encodingdefault}{\sfdefault}{bx}{n}



%% file: Section/0.abstract.tex
\begin{abstract}
We present Unravel, a training-free framework that turns a single image into an interactive 3D scene, where every object is an independent and movable asset. Existing generators either produce one entangled geometry in which no object can be moved, or predict all objects in one feed-forward pass, where a wrong guess cannot be detected or repaired. Unravel instead takes a different approach: it first takes the scene apart, and then puts it back together. Starting from the input image, an agentic harness that iteratively peels the image layer by layer, and adopts point cloud as major modality for spatial relationship recovery. Specifically, a vision-language model plans which objects are fully visible, an image editor removes them and restores the occluded regions. Every edit is verified before being incorporated into the reconstruction. This process gradually turns occluded parts of the scene into direct observation, allowing objects to be reconstructed when they are best exposed rather than hallucinated from the original image alone. The scene is then rebuilt in reverse removal order, so that the supporting objects are naturally placed before the objects they support. Since a single-view image only reveals part of an object, we treat the observed points as a lower bound on its size: an asset may grow to cover its points, but can not shrink away from them. Finally, we physically settle the reconstructed scene to remove floating objects and interpenetrations, followed by an agentic audit that corrects remaining placement errors. Extensive experiments on diverse indoor scenes show that Unravel recovers more objects, places them more faithfully, and produces scenes that are more physically-plausible than existing methods. Our code will be released at \url{https://github.com/SII-Hui/Unravel}.

\end{abstract}

%% file: Section/1.Intro.tex
\section{Introduction}

\begin{figure}[t]
  \centering
  \includegraphics[width=0.85\linewidth]{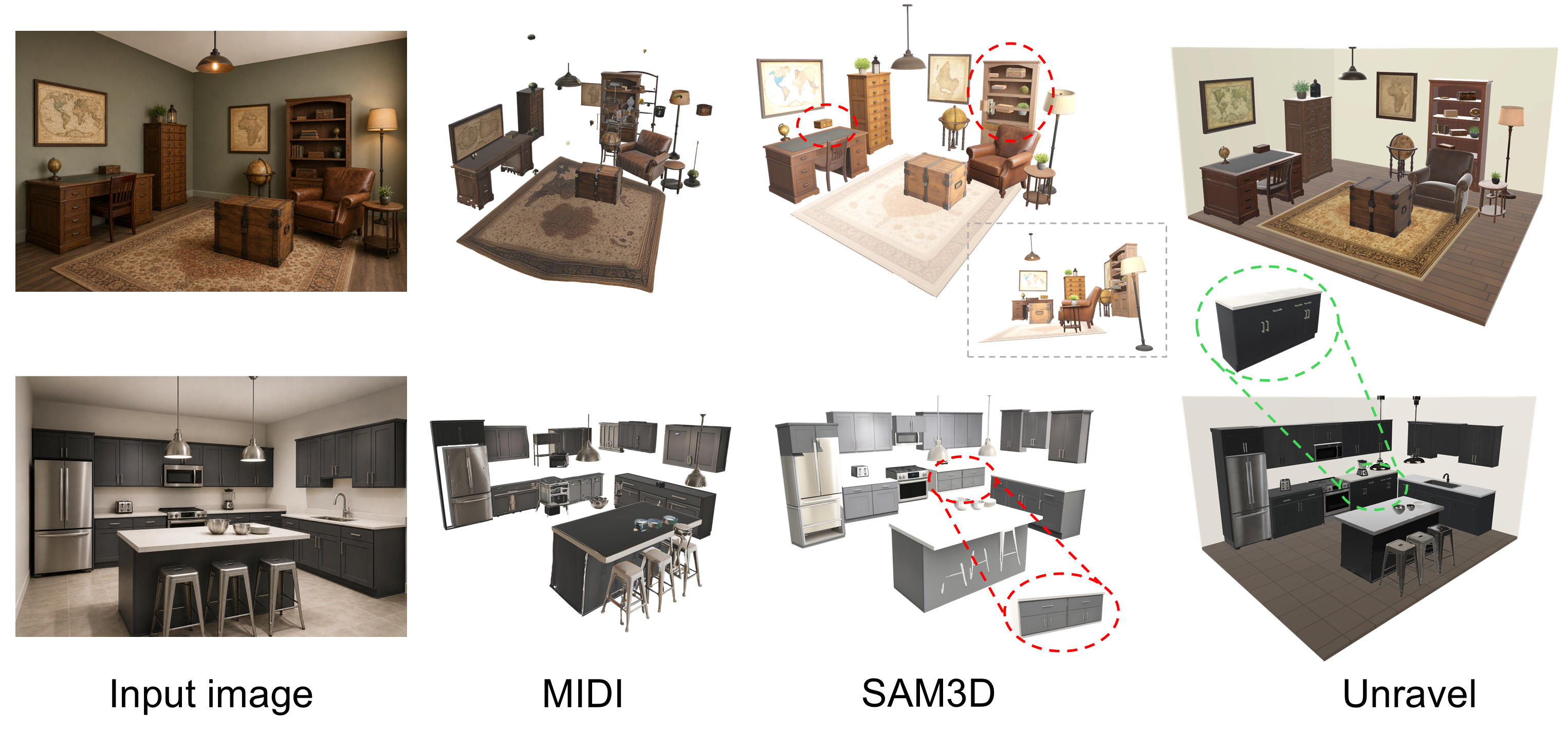}
  \caption{Single-image interactive scene reconstruction. One-shot generators such as MIDI~\citep{huang2025midi} and SAM 3D~\citep{chen2026sam} recover independent assets, but under occlusion they misplace objects and distort the geometry of hidden parts (red highlights). Unravel recovers occluded objects with faithful geometry and placement (green highlights).}
  \label{fig:motivation}
\end{figure}

Embodied agents act on the world around them. They move objects, open containers, and rearrange furniture. Training them in real spaces is too expensive, so they are trained in interactive 3D simulated scenes, where every object can be moved, replaced, or re-arranged~\citep{yang2024holodeck,xia2026sage}. Building these simulated scenes by hand is costly as well, and recovering them from single images offers an appealing alternative. A photograph is cheap to collect, but what we recover from it should go beyond appearance. The scene must stay physically meaningful when its objects are moved.

Many reconstruction methods recover scene geometry without explicit object decomposition~\citep{wang2024dust3r,wang2025vggt,kong2026moge}. They reconstruct visible surfaces but leave occluded regions incomplete and do not directly produce movable assets. Object-centric generators instead predict separate meshes and poses from a cluttered image~\citep{huang2025midi,meng2026scenegen,chen2026sam}. However, learning 3D priors requires costly data and training. Under heavy occlusion, errors in shape completion can distort individual assets and propagate into the layout (Fig.~\ref{fig:motivation}). Because object recovery and placement are coupled, these errors are difficult to isolate and correct.

We present Unravel, a training-free framework that deconstructs a scene before recomposing it (Fig.~\ref{fig:teaser}). At each round, a vision-language model selects exposed objects for removal. An image editor removes them and completes the freed regions, and a verifier checks the edit before it contributes to reconstruction (Fig.~\ref{fig:layerwise}). Occluded content is therefore completed in image space, where intermediate predictions can be inspected and corrected. Each object is captured immediately before removal and converted from its best-observed snapshot into an independent 3D asset.

Iterative removal can provide cleaner inputs for object reconstruction. Our key insight is to use the deconstruction record as a geometric reference for recomposition. We align and fuse the per-round point maps while retaining each point's source round and pixel. Under a fixed camera, source rounds distinguish successive surface observations along the same image ray. Instance masks associate these observations with assets, producing point clusters that anchor their placement. Reversing the removal order also guides assembly from supporting objects to the objects they carry.

Recomposition must respect the incompleteness of these clusters. Unravel registers each asset under one-sided extent constraints on under-observed axes. The observed spans provide lower bounds rather than forcing a complete asset to match a partial surface. Physical settling reduces floating and interpenetration while limiting displacement from the geometric anchors. A final agentic audit inspects scene renderings and refines remaining placement errors. Ablations show improved asset fidelity over direct reconstruction. Qualitative comparisons with feed-forward and agentic baselines show more faithful layouts. Our contributions are threefold.

\begin{itemize}
\item We introduce Unravel, a training-free framework for reconstructing interactive 3D scenes from a single image. Verified layer-wise deconstruction generates independently movable assets from their best-observed snapshots.
\item We develop a recomposition strategy that reuses the deconstruction record. Round-and-pixel provenance anchors each asset to its instance cluster, and reverse removal order guides support-aware assembly. The strategy combines one-sided extent constraints with physical settling.
\item Extensive experiments across diverse indoor scenes demonstrate that Unravel can faithfully reconstruct interactive 3D scenes from a single image.
\end{itemize}

%% file: Section/2.RelatedWork.tex
\section{Related Work}

\subsection{Feed-forward 3D Reconstruction}

Feed-forward models offer an alternative to per-scene optimization, including in pose-free settings. DUSt3R~\citep{wang2024dust3r} regresses aligned pointmaps from uncalibrated image pairs. Its successors support robust matching (MASt3R~\citep{leroy2024grounding}), streaming reconstruction (Spann3R~\citep{wang20253d}, CUT3R~\cite{wang2025continuous}), and thousand-view input (Fast3R~\citep{yang2025fast3r}). VGGT~\citep{wang2025vggt} jointly predicts cameras, depth, pointmaps, and tracks for hundreds of views in one forward pass. For monocular input, Metric3Dv2~\citep{hu2024metric3d} estimates zero-shot metric depth and surface normals. Depth Anything V2~\citep{yang2024depth} provides relative depth. The MoGe series progresses from affine-invariant pointmaps~\citep{wang2025moge} to detailed metric geometry~\citep{wang2026moge,kong2026moge}. These methods recover per-pixel geometry without explicit object decomposition, leaving occluded regions incomplete and objects inseparable.

\subsection{3D Object Asset Generation}

Image-to-3D models generate textured meshes from clean object images. TripoSR~\citep{tochilkin2024triposr} directly infers geometry in one forward pass. Multi-view methods (InstantMesh~\citep{xu2024instantmesh}, LGM~\citep{tang2024lgm}, Unique3D~\citep{wu2024unique3d}) synthesize views before reconstructing geometry. Native 3D latent models (CLAY~\citep{zhang2024clay}, TRELLIS~\citep{xiang2025structured}, Hunyuan3D-2~\citep{zhao2025hunyuan3d}, TripoSG~\citep{li2025triposg}, Direct3D-S2~\citep{wu2026direct3d}) learn directly from asset corpora. TRELLIS.2~\citep{xiang2026native} generates high-resolution textured assets, while NI-Tex~\citep{shan2026ni} produces garment textures from non-isometric image inputs. Part-aware models add component-level structure. PartCrafter~\citep{lin2026partcrafter} jointly generates part meshes from a single image, and HoloPart~\citep{yang2026holopart} completes segmented parts. These models provide strong priors for individual asset generation. Extending them to complete scene generation remains challenging.

\subsection{Interactive 3D Scene Generation}

Object-level scene generation produces independently editable assets rather than fused geometry. Text-driven methods arrange objects through layout diffusion (DiffuScene~\citep{tang2024diffuscene}, InstructScene~\citep{lin2024instructscene}, PhyScene~\citep{yang2024physcene}) or LLM-based planning (Holodeck~\citep{yang2024holodeck}). Agentic systems (SAGE~\citep{xia2026sage}, SceneSmith~\citep{pfaff2026scenesmith}, SceneAssistant~\citep{luo2026sceneassistant}) refine scenes through visual feedback. GPT-6 Astra~\citep{openai_gpt} also supports image understanding, editing, and 3D tool use. Yet semantic instructions alone underspecify the geometry of a reference image. ControlRoom3D~\citep{schult2024controlroom3d} and WonderWorld~\citep{yu2025wonderworld} offer layout-conditioned and single-image scene synthesis, respectively, without targeting independent object editing.

Closer to our setting, MIDI~\citep{huang2025midi}, SceneGen~\citep{meng2026scenegen}, SAM 3D~\citep{chen2026sam}, and CAST~\citep{yao2025cast} infer object meshes and poses from cluttered views. Their 3D priors complete occluded regions but require costly data and training. Training-free pipelines such as Interact3D~\citep{shan2026interact3d} and SimFoundry~\citep{ranawaka2026simfoundry} use modular optimization and self-correction. Precise placement remains challenging in crowded scenes. Seeing Through Clutter~\citep{aguina2026seeing} iteratively removes objects to obtain cleaner reconstruction inputs. Unravel links verified deconstruction to physically constrained recomposition. Round-and-pixel provenance anchors each asset to its reconstructed geometry, and reverse removal order guides support-aware assembly.

%% file: Section/3.Methods.tex
\section{Methods}

\begin{figure}[t]
  \centering
  \includegraphics[width=0.85\linewidth]{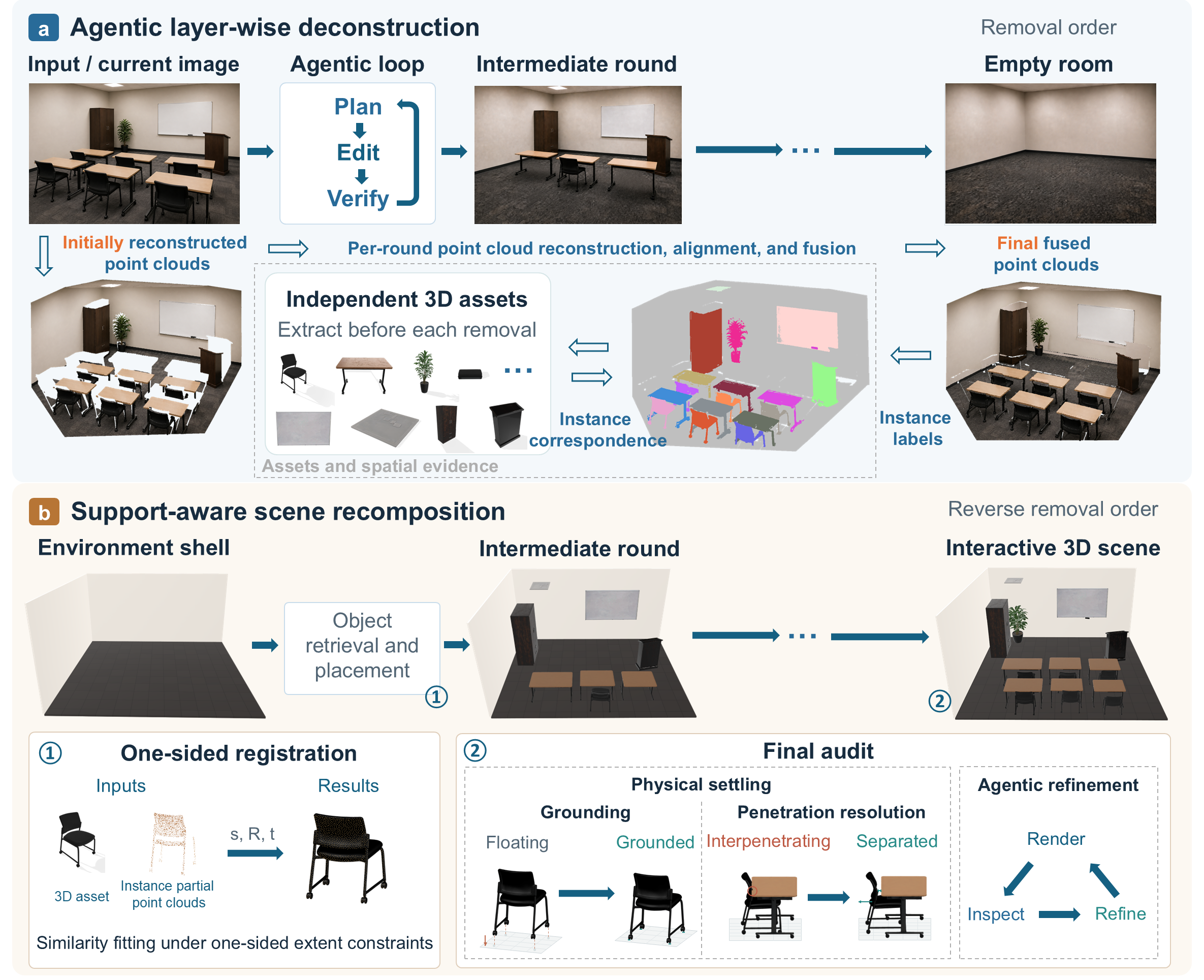}
  \caption{Overview of Unravel. (a) Agentic layer-wise deconstruction. An agentic loop plans, edits, and verifies removals until the room is empty. Independent 3D assets are extracted before removal. Per-round reconstructions are aligned and fused into an instance-labeled point cloud. (b) Support-aware scene recomposition. Assets return to the environment shell in reverse removal order. One-sided extent constraints guide registration to their point clusters. Physical settling and a final audit refine placement.}
  \label{fig:overview}
\end{figure}

Unravel deconstructs the scene before recomposing it (Fig.~\ref{fig:overview}). The deconstruction stage (Section~\ref{subsection:AgenticLayerWise}) removes objects layer by layer and extracts independent assets from their best-observed snapshots. Per-round reconstructions are aligned and fused into a point cloud that retains each object's location. Recomposition (Section~\ref{subsection:SupportAwareRecomposition}) places assets in reverse removal order. Each asset is registered to its point cluster, and physical settling refines the assembled scene.

\subsection{Problem Formulation}
\label{subsection:ProblemFormulation}

Given a single image $I$ of an indoor scene, our goal is to recover an interactive 3D scene $\mathcal{S}$ consisting of an environment shell and a set of independently placed assets,
\begin{equation*}
    \mathcal{S} = \big(\mathcal{E},\ \{(a_i, T_i)\}_{i=1}^{N}\big).
\end{equation*}
The shell $\mathcal{E}$ comprises the floor and the walls. Each $a_i$ is an independent 3D asset placed by $T_i = (\mathbf{s}_i, R_i, t_i)$: a rotation $R_i \in SO(3)$, a translation $t_i \in \mathbb{R}^3$, and a scale $\mathbf{s}_i \in \mathbb{R}_{+}^{3}$. The scale is resolved axis by axis in the asset's upright frame, since a single view bounds the extent of each axis only from below (Section~\ref{subsubsection:OneSidedRegistration}). We additionally record which asset or shell surface each object rests on, so that objects can be moved, replaced, or re-arranged, and the scene re-settled into a physically valid state after any edit.

\subsection{Agentic Layer-Wise Deconstruction}
\label{subsection:AgenticLayerWise}

We use 2D foundation models to plan, edit, and verify layer-wise removals without task-specific training. Occluded content is inferred through verified image completion before 3D reconstruction. With the camera fixed, we align each reconstruction to the first round and fuse them into a point cloud. This cloud provides the primary geometric reference for recovering spatial relationships.

\paragraph{The Deconstruction Loop}
\label{subsubsection:DeconstructionLoop}


\begin{figure}[t]
  \centering
  \includegraphics[width=0.9\linewidth]{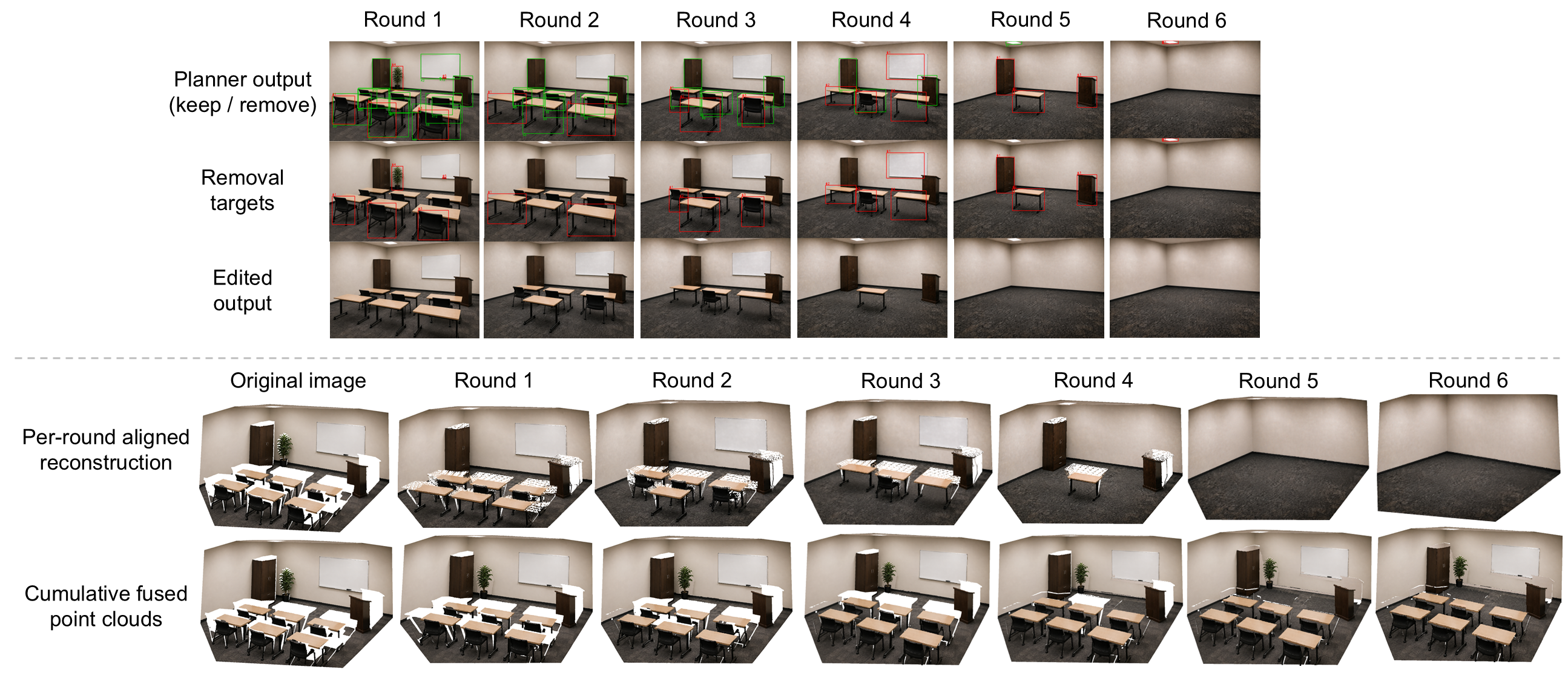}
  \caption{Layer-wise deconstruction and point cloud fusion. The upper block shows planner outputs, removal targets, and edited images across six rounds. Red boxes mark targets, and green boxes mark protected objects. The lower block shows aligned per-round reconstructions and cumulative fused point clouds, starting with the original-image reconstruction. Fusion retains removed-object geometry and adds surfaces reconstructed from edited images.}
  \label{fig:layerwise}
\end{figure}


The loop uses a VLM planner and verifier with an image editor (Fig.~\ref{fig:layerwise}, upper block). The planner selects fully visible objects for removal and marks the rest for protection. Their bounding boxes are drawn on a plan image. Walls and the floor are preserved, and contents are removed before containers. Guided by the plan, the editor removes targets from $I_k$ and inpaints the freed regions to produce $I_{k+1}$. These edits provide more complete object views for later rounds. The verifier checks for missed targets, unintended removals, and fabricated content before an edit is committed. Changed pixels between $I_k$ and the verified $I_{k+1}$ form the edit mask $E_k$. The loop repeats until only the empty room remains.


We extract each asset immediately before removal from its best-observed snapshot. A segmentation model~\citep{carion2026sam} operates inside its bounding box to obtain the instance mask $S_i$, improving precision over full-image segmentation. Appendix~\ref{appendix:boxes} details box localization. We store the snapshot and semantics. gpt-image-2~\citep{openai_gpt} converts the snapshot to a canonical image, which an image-to-3D model~\citep{xiang2026native} turns into a textured mesh. The asset index records the mesh, mask, semantics, box, and removal round for recomposition.

\paragraph{Recovering Scene Geometry Across Rounds}
\label{subsubsection:driftcontrol}


The key challenge in this loop is drift. Each round rebuilds the scene from a different edited image. Without explicit control, errors in scale and pose can misalign surfaces across rounds and distort the fused geometry.


We fix camera intrinsics across rounds. The monocular estimator MoGe3~\citep{kong2026moge} predicts a point map $P_k \in \mathbb{R}^{H \times W \times 3}$ and per-pixel confidence $c_k(p) \in [0, 1]$ from $I_k$. The validity mask $M_k = \{p \mid c_k(p) > \tau\}$ retains pixels above a fixed threshold $\tau$. We also exclude depth discontinuities, where estimates may mix surfaces. Each later reconstruction is aligned directly to the first with a similarity transformation.

We estimate this transformation using only stable pixels, which we call anchors. A pixel serves as an anchor only if its geometry is valid in both the first and the current round and it has never been edited in earlier rounds, so that it depicts the same surface in both rounds. Formally, the anchor set is $\mathcal{A}_k = \{p \mid M_k(p) \wedge M_1(p) \wedge p \notin \bigcup_{j<k} E_j\}$, and we minimize
\begin{equation*}
    \min_{s,R,t}\ \sum_{p \in \mathcal{A}_k} \big\| P_1(p) - \big( sR\,P_k(p) + t \big) \big\|^2.
\end{equation*}
We solve this objective in closed form~\citep{umeyama1991least}. Some anchors nevertheless carry inconsistent geometry across rounds, mostly near editing boundaries and shadow residues. We treat anchors with large residuals after a first fit as outliers, discard them, and refit on the remaining anchors.

The fused cloud accumulates only new information. Unedited pixels keep their existing points. Pixels edited in round $k-1$ contribute the surfaces they now reveal,

\begin{equation*}
    \mathrm{PC} \leftarrow \mathrm{PC} \cup \big\{ sR\,P_k(p) + t \mid p \in E_{k-1} \cap M_k \big\},
\end{equation*}


\noindent where $E_{k-1}$ is the previous edit mask and $M_k$ is the current validity mask. Edit masks select each point's source round, while validity masks filter unreliable estimates. The lower block of Fig.~\ref{fig:layerwise} shows how fusion preserves removed-object geometry and adds surfaces reconstructed from completed regions. Alignment places these points in a shared frame for asset placement.

\paragraph{Point Attribution and Instance Clusters}
\label{subsubsection:PointAttribution}

Labeling a single-view cloud is normally ambiguous: a pixel admits one label, while the surfaces stacked behind it belong to different objects. Deconstruction dissolves this ambiguity. The camera is locked, so a pixel $p$ is one fixed ray, and every round observes this ray at a different depth. We call the points added to the cloud at round $k$ the layer of round k, with the layer 1 the whole first reconstruction. A pixel now holds one point per layer, and we label points rather than pixels.

For the point $x$ of layer $k$ at pixel $p$, the candidates are the objects $a_i$ whose instance masks $S_i$ cover $p$ and that are still present in round $k$, namely, whose removal round satisfies $r_i \ge k$. They queue along the ray in the order of removal, and we label $x$ by the frontmost object. If nothing in the queue survives to round $k$, the ray shows the room, and we label $x$ by $0$, the room. Because a label depends only on its own layer and on the fixed masks, later removals never change labels already assigned. Points labeled $i$ form the cluster $C_i$ of asset $a_i$, and points labeled $0$ form the room structure, from which the environment shell $\mathcal{E}$ is fitted. The removal sequence provides the depth ordering along each ray. Masks come from each object's best-observed image and may include content completed by earlier edits.

\subsection{Support-aware Scene Recomposition}
\label{subsection:SupportAwareRecomposition}

Deconstruction records each object's removal round and geometric observations. We use both to recompose the scene (Fig.~\ref{fig:overview}(b)). Removal rounds guide support-aware assembly. Point provenance anchors registration and physical settling.

\paragraph{From Occlusion Order to Support Order}
\label{subsubsection:OcclusionToSupport}

The deconstruction planner removes contents before their containers, since a container becomes fully visible only after its contents leave. For a support relation $i \rightarrow j$, where object $i$ rests on object $j$, the removal order therefore satisfies $r_i < r_j$. We assemble the scene in descending removal round, starting from the empty shell. For every support relation $i \rightarrow j$, asset $j$ is then placed before asset $i$. Every object arrives when its support is already in place, so placement always has a physical reference. Large supports from later rounds enter first, and small foreground objects from early rounds enter last.

The fused cloud provides the placement evidence. By the attribution rule of Section~\ref{subsubsection:PointAttribution}, every point belongs to one asset, and the points of asset $a_i$ form its cluster $C_i$. The remaining points belong to the room structure, and fitting them gives the environment shell $\mathcal{E}$.

\paragraph{One-Sided Registration under Physical Constraints}
\label{subsubsection:OneSidedRegistration}

Each generated asset is complete, while its cluster $C_i$ covers only surfaces reconstructed from the original and edited images. Registration must fit the asset to its cluster without shrinking below what was observed. Let $X_i$ be surface samples from asset $a_i$. We register them to $C_i$ with a scaled rigid transformation $T_i$. Candidate transformations are scored by a bidirectional median Chamfer distance~\citep{fan2017point},
\begin{equation*}
    D_i(T) = \frac{1}{2}\left[\operatorname{med}_{x \in T(X_i)} \min_{y \in C_i}\|x-y\|_2 + \operatorname{med}_{y \in C_i} \min_{x \in T(X_i)}\|y-x\|_2\right],
\end{equation*}
and the best candidate is refined by truncated similarity ICP. The first term measures asset-to-cluster fit, and the second measures coverage of the cluster. Median distances reduce sensitivity to neighboring-object fragments and spurious generated parts. Truncated ICP similarly retains reliable correspondences at each step to limit distortion from partial observations.


A single view records only the front of an object, so a cluster provides a lower bound on the object's size rather than the exact size. We therefore compare the asset and its cluster along the three axes of the object's upright frame. The height axis points up along gravity. The other two axes lie in the horizontal plane. The depth axis runs front to back, closer to the viewing direction, while the width axis runs across the view. Occlusion hides the object's back, so the depth axis is always under-observed. The width or height axis may also be under-observed when the object is cut off by the image border or largely occluded along that direction. These axes form $\mathcal{U}_i$. Registration solves
\begin{equation*}
    T_i^* = \arg\min_T D_i(T) \quad \text{s.t.} \quad \operatorname{span}_u\big(T(X_i)\big) \geq \operatorname{span}_u(C_i), \quad \forall u \in \mathcal{U}_i,
\end{equation*}
where $\operatorname{span}_u(\cdot)$ measures how far a point set extends along axis $u$. In words, along an under-observed axis the asset must reach at least as far as the observed points. Partial observation can therefore grow an asset, but never shrink it.

In practice the transform stays close to a similarity map. It combines a global scale with bounded per-axis adjustments in the object's frame, and never becomes a free affine map. Appendix~\ref{appendix:registration} gives the exact update rules.

\paragraph{Physical Settling and Final Audit}
\label{subsubsection:PhysicalSettling}

Registration places every asset close to where it was seen, but it does not guarantee a plausible scene. Objects may float above the floor or sink into their neighbors. We correct these defects with small translations. To decide which objects may move, we infer the support graph from semantic relations and local support geometry, and classify each object as mounted or movable. Mounted objects, such as pictures and pendant lights, are attached to walls or the ceiling and stay fixed. Writing the settled asset as $a_i'=T_i(X_i)+\delta_i$, we formulate settling as
\begin{equation*}
    \min_{\{\delta_i\}_{i \in \mathcal{D}}}\sum_{i \in \mathcal{D}}\|\delta_i\|_2^2 \quad \text{s.t.}\quad \operatorname{supported}(a_i')=1,\quad \phi(a_i',a_j')\geq 0\ \ \forall i<j,\quad \delta_i\in\mathcal{B}_i,
\end{equation*}
where $\mathcal{D}$ indexes the movable objects during settling. The objective asks for the smallest total movement, and the constraints state what a plausible scene must satisfy.

Every movable object must be supported, with its bottom resting on the floor or the top of another object. The clearance $\phi(a_i', a_j')$ between two assets turns negative when they overlap, so the second constraint prevents interpenetration. The budget $\mathcal{B}_i$ keeps corrections faithful to the observation, so an asset aligned with its cluster may move only slightly, and a misplaced asset may move toward the cluster but never further away. Appendix~\ref{appendix:registration} gives the exact definitions. Settling lowers floating objects onto their supports and separates interpenetrating pairs. A final agentic audit renders the settled scene and checks for remaining placement errors. Each detected error triggers a local correction followed by another settling pass. The audit repeats until no further error is detected.

%% file: Section/4.Experiments.tex
\section{Experiments}

\subsection{Experimental Setup}
\label{subsection:ExperimentalSetup}

\noindent\textbf{Models and Evaluation Setup.}
We compare Unravel with feed-forward object and pose predictors and agentic baselines based on iterative reasoning and tool use. We use GPT-5.5~\citep{openai_gpt} as the vision-language model for per-round planning, semantic localization, and edit verification. The image editor is gpt-image-2~\citep{openai_gpt}, which also produces the canonical object images. SAM 3~\citep{carion2026sam} extracts instance masks inside the bounding boxes provided by the planner. MoGe-3~\citep{kong2026moge} reconstructs the per-round point maps, and TRELLIS.2~\citep{xiang2026native} turns the canonical images into textured assets. GPT-6 Astra (Light)~\citep{openai_gpt} performs the final agentic audit on the rendered scene. The agentic baseline uses the stronger Astra (Ultra) variant. Calls to GPT-6 Astra are through the official web interface. Every component is used off the shelf, and none is fine-tuned on our data. The framework is training-free.

\noindent\textbf{Baseline Methods.}
Among feed-forward baselines, MIDI~\citep{huang2025midi} generates object instances through multi-instance diffusion. SAM 3D~\citep{chen2026sam} predicts object geometry, texture, and layout from the input image. Both directly recover objects and their arrangement, so we evaluate recovery and placement together. For the agentic comparison, we select GPT-6 Astra (Ultra)~\citep{openai_gpt} for its strong visual understanding and reasoning. We evaluate two settings that differ in mesh source. In the direct setting, Astra generates meshes and their arrangement from the input image. In the shared-asset setting, it receives our 3D assets and 2D segmentation masks (Section~\ref{subsection:AgenticLayerWise}), along with the input image. It estimates only each asset's scale and pose. We denote this variant GPT-6 Astra (Ultra) + TRELLIS2 (Fig.~\ref{fig:Baselines}). Astra generates coarser meshes than image-to-3D models. Sharing assets therefore isolates placement from mesh quality. This recomposition ablation compares Astra's agentic strategy with our point-cluster registration and physical settling. We also tested Seeing Through Clutter~\citep{aguina2026seeing}. The evaluated pipeline lacks an explicit edit-verification harness. Our tests on complex scenes showed substantial hallucinations and unstable editing. Figure~\ref{fig:seeing_comparison} in Appendix~\ref{appendix:results} presents one comparison.

\begin{table}[t]
  \centering
  \caption{Ablation on five scenes. Area ratio and UCR summarize coverage for objects removed after the first round, using Unravel's masks and point clusters as references. UCR (under-coverage rate) denotes the percentage with coverage below 70\%. VLM and Human report mean fidelity scores (1 to 10) over all 160 objects. Point counts are scene averages.}
  \label{tab:ablation}
  \small
  \setlength{\tabcolsep}{3pt}
  \renewcommand{\arraystretch}{1.0}
  \resizebox{0.95\linewidth}{!}{%
    \begin{tabular*}{\linewidth}{@{\extracolsep{\fill}}lcccccc@{}}
      \toprule
       & \multicolumn{2}{c}{2D masks}
       & \multicolumn{2}{c}{3D assets $\uparrow$}
       & \multicolumn{2}{c}{Scene geometry} \\
      \cmidrule(lr){2-3}\cmidrule(lr){4-5}\cmidrule(l){6-7}
      Method & Area ratio $\uparrow$ & UCR (\%) $\downarrow$ & VLM & Human & Points (M) $\uparrow$ & UCR (\%) $\downarrow$ \\
      \midrule
      Direct reconstruction & 0.772 & 11.3 & 4.7 & 4.9 & 1.53 & 10.1 \\
      \textbf{Unravel} & \textbf{1.000} & \textbf{0.0} & \textbf{8.2} & \textbf{8.6} & \textbf{2.81} & \textbf{0.0} \\
      \bottomrule
    \end{tabular*}%
  }
\end{table}

\subsection{Qualitative Comparisons}
\label{subsection:QualitativeComparisons}

\begin{figure}[t]
  \centering
  \includegraphics[width=0.92\linewidth]{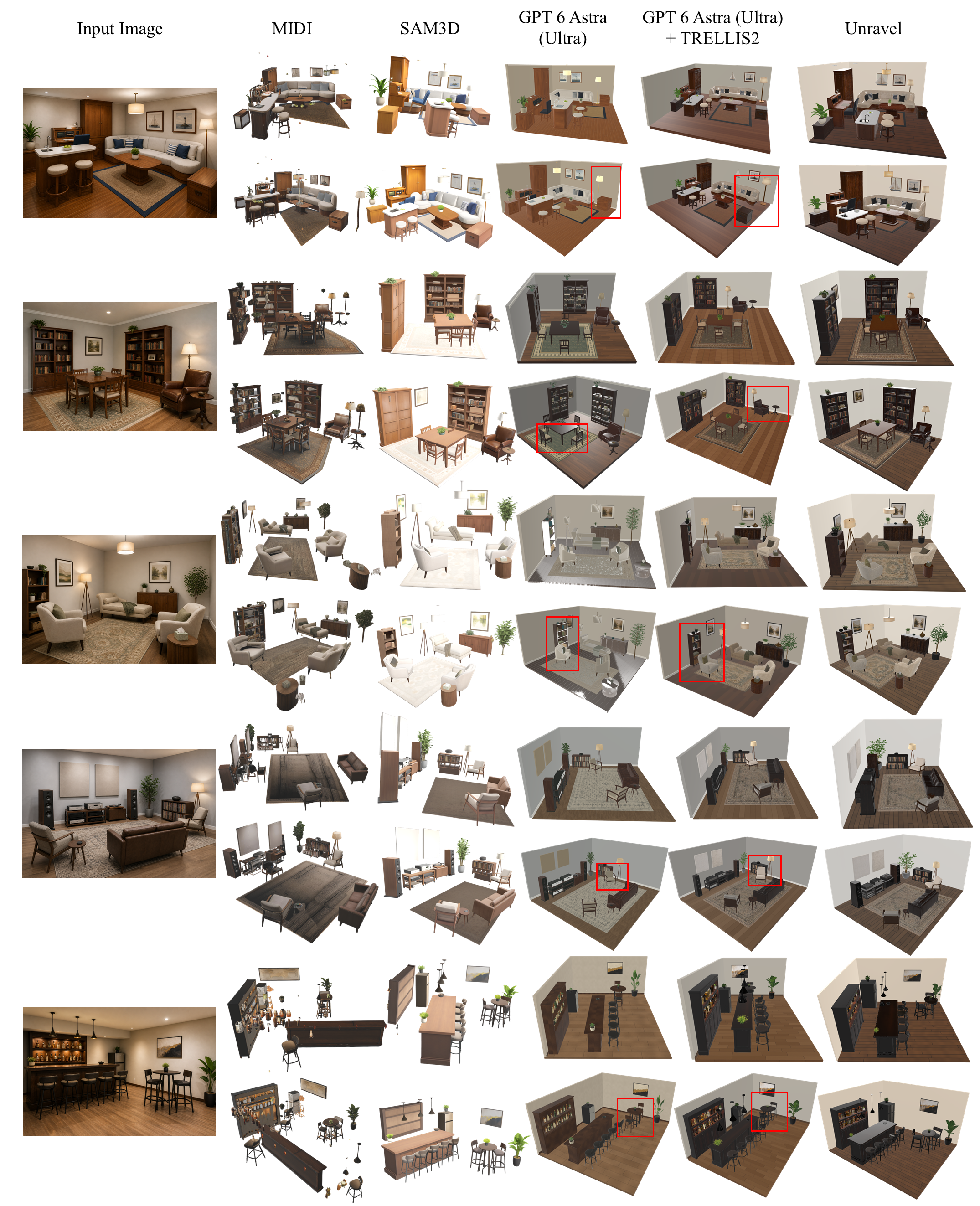}
  \caption{Qualitative comparison of single-image scene reconstruction. Each input is followed by two views of each reconstructed scene. The two Astra variants separate direct scene generation from placement with shared assets. GPT-6 Astra (Ultra) + TRELLIS2 receives the same generated meshes and corresponding 2D instance masks used by Unravel. Red boxes highlight placement discrepancies in the agentic results.}
  \label{fig:Baselines}
\end{figure}

Figure~\ref{fig:Baselines} compares the four baselines on multiple scenes, each shown from two views. MIDI relies on priors learned from limited 3D data, so heavily occluded objects often suffer severe generation errors. SAM 3D produces reasonable assets in most cases, but its textures tend to drift. Its asset quality falls short of TRELLIS.2, a dedicated image-to-3D model, and its training is expensive. Its placement estimates are likewise imprecise, which easily introduces floating and interpenetration. The direct Astra variant captures broad spatial relations, yet its self-generated meshes are visibly coarser. Sharing our assets removes this gap, yet the red boxes still reveal placement discrepancies. Since the meshes are shared, these errors come from placement rather than asset quality. Across these scenes, Unravel preserves the input layout most closely. Appendix~\ref{appendix:results} reports more reconstructions with per-round edits, extracted assets, and multi-view renderings. Figure~\ref{fig:real_results} shows reconstructions from six real indoor images.

\subsection{Ablation Study}
\label{subsection:Ablation}
To evaluate the deconstruction loop, we compare Unravel with direct reconstruction on five scenes containing 160 objects. Both use SAM 3 for segmentation, TRELLIS.2 for asset generation, and MoGe3 for point cloud reconstruction. Direct reconstruction uses only the input image. Unravel extracts each object's mask when it is ready for removal and fuses the point clouds from successive rounds. We compare the resulting 2D masks, 3D assets, and scene geometry in Table~\ref{tab:ablation}.

For 2D masks, we report each object's direct reconstruction mask area as a percentage of its Unravel counterpart. Objects removed after the first round account for roughly one third of the test set. Their mean area ratio is 77.2\%, with means of 85.0\% for round 2 and 60.4\% for round 4. Across all objects, 5.6\% have ratios below 50\%, with a minimum of 23.8\%. Larger masks give TRELLIS.2 more visual detail for asset generation. To assess the resulting 3D assets, a vision-language model and human raters score each mesh against its reference image. Scores range from 1 to 10, with higher values indicating greater fidelity. Appendix~\ref{appendix:scoring} details the protocol and examples. For scene geometry, fusing four to eight rounds increases the mean point count per scene from 1.53 million to 2.81 million. We measure the fraction of points in Unravel's fused object clusters covered by direct reconstruction. Mean coverage is 76.9\% for objects removed after the first round. Across evaluated objects, 10.1\% have coverage below 70\% and 3.8\% below 50\%. The minimum is 11.9\%. These measurements quantify the additional geometry available for asset placement after deconstruction.

%% file: Section/5.Conclusions.tex
\section{Limitations and Future Work}

Unravel is limited by current image editing models. In highly cluttered scenes, edits may alter the appearance or position of remaining objects. Such drift can accumulate across rounds and disrupt point-cloud alignment and asset placement. Our verification improves editing reliability, but minor errors may persist. Future work could combine stronger image editing models with geometric feedback to better preserve object identity and spatial relationships across rounds. Uncertainty-aware checks could trigger targeted retries before small errors accumulate.